\RequirePackage{fix-cm}
\documentclass{svjour3}                     

\smartqed  
\usepackage[utf8]{inputenc}
\usepackage{amssymb,amsfonts,bm,amsmath} 
\usepackage{cite}

\usepackage{graphicx}
\usepackage{float}
\usepackage{url}

\begin{document}

\title{Global convergence of Negative Correlation Extreme Learning Machine}
\author{Carlos Perales-Gonz\'alez}
\authorrunning{C. Perales-Gonz\'alez}

\institute{Universidad Loyola Andalucía,\\
	Sevilla, Spain. \\
	Tel.: +34 955641600 \\
	\email{cperales@uloyola.es}}

\date{This is a preprint of an article published in Neural Processing Letters. \\
	 The final authenticated version is
available online at: \url{https://doi.org/10.1007/s11063-021-10492-z}}

\maketitle

\begin{abstract}
Ensemble approaches introduced in the Extreme Learning Machine (ELM) literature mainly come from methods that relies on data sampling procedures, under the assumption that the training data are heterogeneously enough to set up diverse base learners. To overcome this assumption, it was proposed an ELM ensemble method based on the Negative Correlation Learning (NCL) framework, called Negative Correlation Extreme Learning Machine (NCELM). This model works in two stages: i) different ELMs are generated as base learners with random weights in the hidden layer, and ii) a NCL penalty term with the information of the ensemble prediction is introduced in each ELM minimization problem, updating the base learners, iii) second step is iterated until the ensemble converges.

Although this NCL ensemble method was validated by an experimental study with multiple benchmark datasets, no information was given on the conditions about this convergence. This paper mathematically presents the sufficient conditions to guarantee the global convergence of NCELM. The update of the ensemble in each iteration is defined as a contraction mapping function, and through Banach theorem, global convergence of the ensemble is proved.

\end{abstract}

\section{Introduction}
\label{sec:intro}

Over the years, Extreme Learning Machine (ELM) \cite{Huang2012} has become a competitive algorithm for diverse machine learning tasks: time series prediction \cite{ren2019classification}, speech recognition \cite{xu2019connecting}, deep learning architectures \cite{chang2018deep,chaturvedi2018bayesian}, \ldots. Both the Single-Hidden-Layer Feedforward Network (SLFN) and the kernel trick versions \cite{Huang2012} are widely used in supervised machine learning problems, due mainly to its low computational burden and its powerful nonlinear mapping capability. The neural network version of the ELM framework relies on the randomness of the weights between the input and the hidden layer, to speed the training stage while keeping competitive performance results \cite{li2020extreme}.

Ensemble learning, also known as committee-based learning \cite{Zhou2012,Kuncheva2003}, has attracted much interest in the machine learning community \cite{Zhou2012} and has been applied widely in many real-world tasks such as object detection, object recognition, and object tracking \cite{Girshick2014,Wang2012,zhou2014ensemble,Ykhlef2017}. The main characteristic of these methodologies lies in the training data to generate diversity among the base learners. The ensemble methods can be separated whether they promote the diversity implicitly (for example, using data sampling methods, such as Bagging \cite{Breiman1996} and Boosting \cite{Freund1995}) or explicitly (introducing parameter diversity terms, such as Negative Correlation Learning Framework \cite{masoudnia2012incorporation,HuanhuanChen2009}). In this context, Bagging and Boosting are the most common approaches \cite{domingos1997does,wyner2017explaining}, although the convergence of these ensemble methods is not always assured \cite{Rudin2004,Mukherjee2013}.

Negative Correlation Learning is a framework, originally designed for neural network ensemble, that introduces the promotion of the diversity among the base learners as another term to optimize in the training stage of the model \cite{HuanhuanChen2009}. This ensemble learning method has been applied to multi-class problems \cite{Wang2010}, deep learning tasks \cite{shi2018crowd} and semi-supervised machine learning problems \cite{Chen2018}. In the Extreme Learning Machine community, Negative Correlation Extreme Learning Machine was introduced by adding to the $L_2$ regularized ELM \cite{HuanhuanChen2009} the diversity term directly in the loss function \cite{perales2020negative}. This allows managing de diversity along with the regularization and the error. However, this method relies on the convergence of the ensemble, and it was not clarified in the original paper.

In this paper, training conditions for convergence are presented and discussed. The training stage of Negative Correlation Extreme Learning Machine (NCELM) is reformulated as a fixed-point iteration, and the solution of each step can be represented as a contraction mapping. Using Banach Theorem, this contraction mapping implies there is a convergence, and the ensemble method is stable.

The manuscript is organized as follows: Extreme Learning Machine for classification problems and the ensemble method Negative Correlation Extreme Learning machine are explained in Section \ref{sec:ncelm}. Conditions about convergence are studied in Section \ref{sec:convergence}, and discussion about hyper-parameter boundaries and graphic examples are in Section \ref{sec:discussion}. Conclusions are on the final segment of the article, Section \ref{sec:conclusions}.

\section{Negative Correlation Extreme Learning Machine and its formulation}
\label{sec:ncelm}

\subsection{Extreme Learning Machine as base learner}
\label{subsec:elm}

For a classification problem, training data could be represented as $\{\boldsymbol{x}_n, \boldsymbol{y}_n\}_{n = 1}^N$, where

\begin{itemize}
\item $\boldsymbol{x} \in \mathbb{R}^K$ is the vector of features of the $n$-th training pattern,
\item $K$ is the dimension of the input features,
\item $\boldsymbol{y}_n \in \mathbb{R}^J$ is the target of the $n$-th training pattern, 1-of-J encoded (all elements of the vector are 0 except the corresponding to the label of the pattern, which is 1),
\item $J$ is the number of classes.
\end{itemize}

Following this notation, the output function of the Extreme Learning Machine classifier \cite{Huang2012} is $\boldsymbol{f} ( \boldsymbol{x}) = \left( \mathbf{f}_1, \ldots, \mathbf{f}_j, \ldots, \mathbf{f}_J  \right)$, where each $\mathbf{f}_j : \mathbb{R}^K \rightarrow \mathbb{R}$ is

\begin{equation}
\label{eq:f_j}
\mathbf{f}_j (\mathbf{x})=\mathbf{h}'(\mathbf{x})\boldsymbol{\beta}_j,
\end{equation}

where $\boldsymbol{h} : \mathbb{R}^K \rightarrow \mathbb{R}^D$ is the hidden layer output. The predicted class corresponds to the vector component with highest value,

\begin{equation}
	\arg\max_{j=1,\ldots,J} \mathbf{f}_j (\mathbf{x}).
\end{equation}

The ELM model estimates the coefficient vectors ${\boldsymbol{\beta}_j} \in \mathbb R^{D}$, where $D$ is the number of nodes in the hidden layer, that minimizes the following equation:

\begin{equation}
\label{eq:elm}
\min_{\boldsymbol{\beta}_j \in \mathbb R ^{D}}\  \left(\Vert \boldsymbol{\beta}_j \Vert^2+C\Vert\mathbf{H}\boldsymbol{\beta}_j-\mathbf{Y}_j \Vert^2 \right), \: j= 1, \ldots, J,
\end{equation}

where

\begin{itemize}
	\item $\mathbf{H} = \left(\mathbf{h}' \left(\mathbf{x}_{1}\right), \ldots, \mathbf{h}' \left(\mathbf{x}_{N}\right)\right) \in\mathbb{R}^{N \times D}$ is the output of the hidden layer for the training patterns,
	\item $\mathbf{Y} = (\mathbf{Y}_{1},\ldots, \mathbf{Y}_{J}) = \left( {\begin{array}{c}
		\boldsymbol{y}_{1}'  \\
		\vdots  \\
		\boldsymbol{y}_{N}'\\
		\end{array} } \right)\in \mathbb R^{N\times J}$ is the matrix with the desired targets
	\item $\mathbf{Y}_{j}$ is the $j$-th column of the $\mathbf{Y}$ matrix.
\end{itemize}

Because Eq. \eqref{eq:elm} is a convex minimization problem, the minimum of Eq. \eqref{eq:elm} can be found by deriving respect to $\boldsymbol{\beta}_j$ and equaling to 0,

\begin{equation}\label{eq:beta}
{\boldsymbol{\beta}}_j = \left( \frac{\mathbf{I}}{C} +\mathbf{H}'\mathbf{H}\right)^{-1}\mathbf{H}'\mathbf{Y}_j.
\end{equation}

\subsection{Negative Correlation Extreme Learning Machine}
\label{subsec:ncelm}

Negative Correlation Extreme Learning Machine model \cite{perales2020negative} is an ensemble of $S$ base learners, and each $s$-th base learner is an ELM, $s=1, \ldots, S$, where $S$ is the number of base classifiers. The result output of a testing instance $\boldsymbol{x} \in \mathbb{R}^K$ is defined as the average of their outputs,

\begin{equation}
\label{eq:ensemble_output}
\boldsymbol{f}_j (\boldsymbol{x}) = \frac{1}{S} \sum_{s=1}^S \mathbf{f}_j^{(s)} (\mathbf{x}) = \frac{1}{S} \sum_{s=1}^S  \mathbf{h}^{(s)'}(\mathbf{x}) \boldsymbol{\beta}_j^{(s)}.
\end{equation}

In the Negative Correlation Learning proposal for ELM framework \cite{perales2020negative}, minimization problem for each $s$-th base learner is similar to Eq. \eqref{eq:elm}, but the diversity among the outputs of the individual $\mathbf{f}_j^{(s)}$, $s = 1, \ldots, S$ and the final ensemble $\boldsymbol{f}_j$ is introduced as a penalization, with $\lambda \in (0, + \infty)$ as a problem-dependent parameter that controls the diversity. The minimization problem for each $s$-th base learner is

\begin{equation}
\label{eq:error_function}
\min_{\boldsymbol{\beta}_j^{(s)} \in \mathbb R^{D\times J}}\ \left( \Vert \boldsymbol{\beta}_j^{(s)} \Vert^2+C\Vert\mathbf{H}^{(s)}\boldsymbol{\beta}_j^{(s)}-\mathbf{Y}_j \Vert^2 +\lambda \left \langle \mathbf{H}^{(s)}\boldsymbol{\beta}_{j}^{(s)}, \boldsymbol{F}_j \right \rangle^2  \right),
\end{equation}

where $\boldsymbol{F}_j$ is the output of the ensemble,
\begin{equation}
\label{eq:F_j}
\boldsymbol{F}_j = \sum_{s'=1}^{S} \boldsymbol{H}^{(s')} \boldsymbol{\beta}_j^{(s')}.
\end{equation}

Because $\boldsymbol{\beta}_j^{(s)}$ appears in $\boldsymbol{F}_j$, the proposed solution for Eq. \eqref{eq:error_function} is to transform the problem in an iterated sequence, with solution of Eq. \eqref{eq:elm} as the first iteration $\boldsymbol{\beta}_{j, (1)}^{(s)}$, for $s=1, \ldots, S$. The output weight matrices in the $r$-th iteration $\boldsymbol{\beta}_{j, (r)}^{(s)}$, $s=1,\ldots,S$, for each individual are obtained from the following optimization problem 

\begin{equation}
\label{eq:error_function_iter}
\min_{\boldsymbol{\beta}_{j, (r)}^{(s)} \in \mathbb R^{D\times J}}\ \left( \Vert \boldsymbol{\beta}_{j, (r)}^{(s)} \Vert^2+C\Vert\mathbf{H}^{(s)}\boldsymbol{\beta}_{j, (r)}^{(s)}-\mathbf{Y}_j \Vert^2 +\lambda\left \langle \mathbf{H}^{(s)} \boldsymbol{\beta}_{j, (r)}^{(s)}, \mathbf{F}_{j, (r-1)}\right \rangle^2  \right),
\end{equation}

where $\boldsymbol{F}_{j, (r - 1)}$ is updated as

\begin{equation}
\label{eq:F_update}
\boldsymbol{F}_{j, (r - 1)} = \frac{1}{S} \sum_{s=1}^{S} \boldsymbol{H}^{(s)} \boldsymbol{\beta}_{j, (r - 1)}^{(s)}.
\end{equation}

As Eq. \eqref{eq:elm}, solution can be obtained for \eqref{eq:error_function_iter} by deriving it and equaling to $0$,

\begin{equation}
\label{eq:solution_iter}
{\boldsymbol{\beta}}_{j, (r)}^{(s)} =  \left( \frac{\mathbf{I}}{C}+\mathbf{H}^{(s)'}\mathbf{H}^{(s)}+\frac{\lambda}{C}\mathbf{H}^{(s)'}\mathbf{F}_{j, (r-1)} \mathbf{F}_{j, (r-1)}^{'} \mathbf{H}^{(s)} \right)^{-1} \mathbf{H}^{(s)'}\mathbf{Y}_j.
\end{equation}

The result ${\boldsymbol{\beta}}_{j, (r)}^{(s)}$ is introduced in $\mathbf{F}_{j, (r)}$ in order to obtain ${\boldsymbol{\beta}}_{j, (r + 1)}^{(s)}$ iteratively. However, the convergence of this iteration $r = 1, \ldots, R$ was not assured in the original paper \cite{perales2020negative}, but it can be proved with Banach fixed-point theorem.

\section{Conditions for the convergence of NCELM}
\label{sec:convergence}

\subsection{Banach fixed-point theorem}
\label{subsec:banach}

As Stephen Banach defined \cite{Banach1922},

\begin{theorem}
Let $(\boldsymbol{X}, d)$ be a non-empty complete metric space with a contraction mapping $T : \boldsymbol{X} \rightarrow \boldsymbol{X}$. Then $T$ admits a unique fixed-point $x^{*}$ in $\boldsymbol{X}$ ($T (x^{*}) = x^{*}$). Furthermore, $x^{*}$ can be found as follows: start with an arbitrary element $x_0 \in \boldsymbol{X}$ and define a sequence ${x_n}: x_n = T(x_{n-1}), \: n > 1$, then $x_n \rightarrow x^{*}$.
\end{theorem}

Let $(\boldsymbol{X}, d)$ a complete metric space, then a map $T : \boldsymbol{X} \rightarrow \boldsymbol{X}$ is called a contraction mapping on $\boldsymbol{X}$ if there exists $q \in [0, 1)$ such that

\begin{equation}
\label{eq:contraction}
d ( \boldsymbol{T} ( x ), \boldsymbol{T} ( y )) \le q d ( x, y ), \: \forall x, y \in \boldsymbol{X}.
\end{equation}

This means that the points, after applying the mapping, are closer than in their original position \cite{Ciesielski2007}. Thus, if NCELM solution in Eq. \eqref{eq:solution_iter} is a contraction mapping on the solutions of ${\boldsymbol{\beta}}_{j, (r)}^{(s)}$, it can be assured that a fixed point exist for NCELM model.

\subsection{Reformulation of NCELM model as a contraction mapping}
\label{subsec:proof}

In order to prove that the iteration of Eq. \eqref{eq:error_function_iter} over $j = 1, \ldots, J$, $s = 1, \ldots, S$ is a fixed-point iteration, the elements of the NCELM model are going to be defined into a metric space $(\mathbb{B}, d)$ with a map $\boldsymbol{T} : \mathbb{B} \rightarrow \mathbb{B}$. Later, it is proved that $\boldsymbol{T}$ is a contraction mapping. An element $\boldsymbol{B} \in \mathbb{B}$ is defined as

\begin{equation} \label{eq:B}
\boldsymbol{B} =
\begin{pmatrix}
\boldsymbol{\beta}_{j}^{(1)} \\
 \vdots \\
\boldsymbol{\beta}_{j}^{(s)} \\
  \vdots \\
\boldsymbol{\beta}_{j}^{(S)} \\
\end{pmatrix},
\end{equation}

thus $\mathbb{B}$ is the subspace that contains the posible solutions of Eq. \eqref{eq:solution_iter}, and it is included in the space $\mathbb{R}^{(D \cdot S) \times 1}$. The output of the ensemble, $\boldsymbol{F}_j$, is then a function of $\boldsymbol{B}$, since it is composed by all the $\boldsymbol{\beta}_j^{(s)}$ by definition in Eq. \eqref{eq:F_j}. Noting this as $\boldsymbol{F}_{\boldsymbol{B}}$, the map $\boldsymbol{T}$ 

\begin{equation} \label{eq:T}
\boldsymbol{T}(\boldsymbol{B}) =
\begin{pmatrix}
{ \left( \frac{\mathbf{I}}{C}+\mathbf{H}^{(1)'}\mathbf{H}^{(1)}+\frac{\lambda}{C}\mathbf{H}^{(1)'} \boldsymbol{F}_{B} \boldsymbol{F}_{B}^{'} \mathbf{H}^{(1)} \right)^{-1} \mathbf{H}^{(1)'}\mathbf{Y}_j} \\
\vdots \\
{ \left( \frac{\mathbf{I}}{C}+\mathbf{H}^{(s)'}\mathbf{H}^{(s)}+\frac{\lambda}{C}\mathbf{H}^{(s)'} \boldsymbol{F}_{B} \boldsymbol{F}_{B}^{'} \mathbf{H}^{(s)} \right)^{-1} \mathbf{H}^{(s)'}\mathbf{Y}_j} \\
\vdots \\
{ \left( \frac{\mathbf{I}}{C}+\mathbf{H}^{(S)'}\mathbf{H}^{(S)}+\frac{\lambda}{C}\mathbf{H}^{(S)'} \boldsymbol{F}_{B} \boldsymbol{F}_{B}^{'} \mathbf{H}^{(S)} \right)^{-1} \mathbf{H}^{(S)'}\mathbf{Y}_j} \\
\end{pmatrix},
\end{equation}

is the applied Equation \eqref{eq:solution_iter} to this point $\boldsymbol{B}$. The map $\boldsymbol{T}$ depends of each classification problem, because of $C$, $\boldsymbol{Y}_j$ and $\boldsymbol{H}^{(s)}, \: s = 1, \ldots, S$ are problem-dependent. Individuals $\boldsymbol{T}^{(s)}$ can be considered,

\begin{equation} \label{eq:T_s}
\boldsymbol{T}(\boldsymbol{B}) =
\begin{pmatrix}
\boldsymbol{T}^{(1)} (\boldsymbol{B}) \\
\vdots \\
\boldsymbol{T}^{(S)} (\boldsymbol{B}) \\
\end{pmatrix}.
\end{equation}

Following this formulation, the NCELM model always starts from the initial point

\begin{equation} \label{eq:B_0}
\boldsymbol{B}_{(0)} =
\begin{pmatrix}
\vec{0} \\
\vdots \\
\vec{0}
\end{pmatrix},
\end{equation}

that leads to $\boldsymbol{F}_{\boldsymbol{B}_{(0)}} = \frac{1}{S} \sum_{s=1}^{S} \boldsymbol{H}^{(s)} \vec{0} = \vec{0}$, thus the first element $\boldsymbol{B}_{(1)}$ in the sequence is

\begin{equation} \label{eq:B_1}
\boldsymbol{B}_{(1)} = \boldsymbol{T} (\boldsymbol{B}_{(0)}) =
\begin{pmatrix}
\left( \frac{\mathbf{I}}{C} +\mathbf{H}^{(1)'}\mathbf{H}^{(1)} \right)^{-1}\mathbf{H}^{(1)'}\mathbf{Y}_j \\
\vdots \\
\left( \frac{\mathbf{I}}{C} +\mathbf{H}^{(s)'}\mathbf{H}^{(s)} \right)^{-1}\mathbf{H}^{(s)'}\mathbf{Y}_j \\
\vdots \\
\left( \frac{\mathbf{I}}{C} +\mathbf{H}^{(S)'}\mathbf{H}^{(S)} \right)^{-1}\mathbf{H}^{(S)'}\mathbf{Y}_j
\end{pmatrix},
\end{equation}

and the problem from Eq.\eqref{eq:error_function_iter} is a sequence $\{ \boldsymbol{B}_{(r)} \} : {\boldsymbol{B}_{(r)} = \boldsymbol{T} (\boldsymbol{B}_{(r-1)})}$, for $r \ge 1$. In the following Section, it is shown that each $\boldsymbol{T}^{(s)}$ is a contraction map, so it is $\boldsymbol{T}$ because Eq. \eqref{eq:T_s}.

\subsection{Definition of distance}
\label{subsec:distance}

For two points from the space $\boldsymbol{U}, \boldsymbol{V} \in \mathbb{B}$, it is defined the distance metric $d(\cdot, \cdot) : \mathbb{R}^{(D \times S) \times 1} \times \mathbb{R}^{(D \times S) \times 1} \rightarrow \mathbb{R}$ as

\begin{equation}
\label{eq:distance}
d (\boldsymbol{U}, \boldsymbol{V}) = \sum_{s=1}^{S} { \lVert \boldsymbol{U}^{(s)} - \boldsymbol{V}^{(s)} \rVert }^2 = \sum_{s=1}^{S} d^{(s)} (\boldsymbol{U}^{(s)}, \boldsymbol{V}^{(s)}),
\end{equation}

where $d^{(s)} (\cdot, \cdot) : \mathbb{R}^{D \times 1} \times \mathbb{R}^{D \times 1} \rightarrow \mathbb{R}$ is the $L_2$ norm power to 2,

\begin{equation}
\label{eq:subdistance}
d^{(s)} (\boldsymbol{U}^{(s)}, \boldsymbol{V}^{(s)}) = {\lVert \boldsymbol{U}^{(s)} - \boldsymbol{V}^{(s)} \rVert }^2.
\end{equation}

The distance after the map $\boldsymbol{T}$ is

\begin{equation}
\label{eq:T_distance}
d ( \boldsymbol{T} (\boldsymbol{U}), \boldsymbol{T} (\boldsymbol{V})) = \sum_{s=1}^{S} { \lVert \boldsymbol{T}^{(s)} (\boldsymbol{U}) - \boldsymbol{T}^{(s)} (\boldsymbol{V}) \rVert }^2 = \sum_{s=1}^{S} d^{(s)} (\boldsymbol{T}^{(s)} (\boldsymbol{U}), \boldsymbol{T}^{(s)} (\boldsymbol{V})).
\end{equation}

so distance $d (\cdot, \cdot)$ is just a sum of $d^{(s)} (\cdot, \cdot)$. It is trivial that if 

$$d^{(s)} (\boldsymbol{T}^{(s)} (\boldsymbol{U}), \boldsymbol{T}^{(s)} (\boldsymbol{V})) \le d^{(s)} (\boldsymbol{U}^{(s)}, \boldsymbol{V}^{(s)}) $$,

then

$$d ( \boldsymbol{T} (\boldsymbol{U}), \boldsymbol{T} (\boldsymbol{V})) < d (\boldsymbol{U}, \boldsymbol{V})$$,

so it is only needed to prove that

\begin{equation}
\label{eq:ineq_distance}
d^{(s)} (\boldsymbol{T}^{(s)} (\boldsymbol{U}), \boldsymbol{T}^{(s)} (\boldsymbol{V}^{(s)})) \le q  {d^{(s)} (\boldsymbol{U}^{(s)}, \boldsymbol{V}^{(s)})} , \: \forall s,  \: q \in [0, 1).
\end{equation}

\subsection{Proof that T is a contraction mapping}
\label{subsec:T}

After computing the training data, the coefficient matrix $\boldsymbol{H}^{(1)}, \ldots, \boldsymbol{H}^{(S)}$ are fixed. If both points $\boldsymbol{U}, \boldsymbol{V}$ are obtained by Eq. \eqref{eq:beta}, $\boldsymbol{U} = \boldsymbol{V}$, and by Eq. \eqref{eq:solution_iter}, $\boldsymbol{T} (\boldsymbol{U}) = \boldsymbol{T} (\boldsymbol{V})$, because both equations give unique solution, and in this case the inequality from Eq. \eqref{eq:ineq_distance} is assured.

Let assume arbitrary $\hat{\boldsymbol{F}}_{\boldsymbol{U}}, \hat{\boldsymbol{F}}_{\boldsymbol{V}}$, initial points from Eq. \eqref{eq:solution_iter} are, 

\begin{eqnarray}
\label{eq:U_j_s}
\boldsymbol{U}^{(s)} \equiv \left( \frac{\mathbf{I}}{C} +\mathbf{H}^{(s)'}\mathbf{H}^{(s)} + \frac{\lambda}{C} \mathbf{H}^{(s)'} \hat{\boldsymbol{F}}_{\boldsymbol{U}} \hat{\boldsymbol{F}}_{\boldsymbol{U}}' \mathbf{H}^{(s)} \right)^{-1} \boldsymbol{H}^{(s)} \boldsymbol{Y}_j, \\
\label{eq:V_j_s}
\boldsymbol{V}^{(s)} \equiv \left( \frac{\mathbf{I}}{C} +\mathbf{H}^{(s)'}\mathbf{H}^{(s)} + \frac{\lambda}{C} \mathbf{H}^{(s)'} \hat{\boldsymbol{F}}_{\boldsymbol{V}} \hat{\boldsymbol{F}}_{\boldsymbol{V}}' \mathbf{H}^{(s)} \right)^{-1} \boldsymbol{H}^{(s)} \boldsymbol{Y}_j.
\end{eqnarray}

From these $\boldsymbol{U}, \boldsymbol{V}$ new predictions $\boldsymbol{F}$ can be obtained, 
\begin{eqnarray}
\label{eq:F_U}
\boldsymbol{F}_{\boldsymbol{U}} = \frac{1}{S} \sum_{s=1}^{S} \boldsymbol{H}^{(s)} \boldsymbol{U}^{(s)}, \\
\label{eq:F_V}
\boldsymbol{F}_{\boldsymbol{V}} = \frac{1}{S} \sum_{s=1}^{S} \boldsymbol{H}^{(s)} \boldsymbol{V}^{(s)}.
\end{eqnarray}

Note that an example of $\hat{\boldsymbol{F}}_{\boldsymbol{U}}, \hat{\boldsymbol{F}}_{\boldsymbol{V}}$ could be $\hat{\boldsymbol{F}}_{\boldsymbol{U}} = \vec{0}$, $\hat{\boldsymbol{F}}_{\boldsymbol{V}} = \boldsymbol{F}_{\boldsymbol{U}}$.

The application of $\boldsymbol{T}^{(s)}$ would result in 

\begin{eqnarray}
\label{eq:t_U_j_s}
\boldsymbol{T}^{(s)} (\boldsymbol{U}) = \left( \frac{\mathbf{I}}{C} +\mathbf{H}^{(s)'}\mathbf{H}^{(s)} + \frac{\lambda}{C} \mathbf{H}^{(s)'} \boldsymbol{F}_{\boldsymbol{U}} \boldsymbol{F}_{\boldsymbol{U}}' \mathbf{H}^{(s)} \right)^{-1} \boldsymbol{H}^{(s)} \boldsymbol{Y}_j, \\
\label{eq:t_V_j_s}
\boldsymbol{T}^{(s)} (\boldsymbol{V}) = \left( \frac{\mathbf{I}}{C} +\mathbf{H}^{(s)'}\mathbf{H}^{(s)} + \frac{\lambda}{C} \mathbf{H}^{(s)'} \boldsymbol{F}_{\boldsymbol{V}} \boldsymbol{F}_{\boldsymbol{V}}' \mathbf{H}^{(s)}  \right)^{-1} \boldsymbol{H}^{(s)} \boldsymbol{Y}_j.
\end{eqnarray}

In order to apply Woodbury matrix identity \cite{Woodbury1950} in Eq. \eqref{eq:t_U_j_s}, the following matrix are renamed:
\begin{itemize}
	\item $\boldsymbol{A}_{\boldsymbol{U}}^{(s)} = \frac{\mathbf{I}}{C} +\mathbf{H}^{(s)'}\mathbf{H}^{(s)} + \frac{\lambda}{C} \mathbf{H}^{(s)'} \hat{\boldsymbol{F}}_{\boldsymbol{U}} \hat{\boldsymbol{F}}_{\boldsymbol{U}}' \mathbf{H}^{(s)}$,
	\item $\boldsymbol{C} = \frac{\lambda}{C} \boldsymbol{I}$ 
    \item $\boldsymbol{D} = \mathbf{H}^{(s)'} $
	\item $\boldsymbol{E} = ({\boldsymbol{F}}_{\boldsymbol{U}} {\boldsymbol{F}}_{\boldsymbol{U}}' - \hat{\boldsymbol{F}}_{\boldsymbol{U}} \hat{\boldsymbol{F}}_{\boldsymbol{U}}' ) \mathbf{H}^{(s)} = \boldsymbol{\delta}_{{\boldsymbol{U}}} \mathbf{H}^{(s)}$,
\end{itemize}

so the inverse of matrix in Eq. \eqref{eq:t_U_j_s} can be rewritten as

\begin{equation}
\label{eq:sherman-morrison}
\begin{split}
\left(\boldsymbol{A}_{\boldsymbol{U}}^{(s)} + \boldsymbol{D} \boldsymbol{C} \boldsymbol{E}  \right)^{-1} & =  \boldsymbol{A}_{\boldsymbol{U}}^{(s), -1} - \boldsymbol{A}_{\boldsymbol{U}}^{(s), -1} \boldsymbol{H}^{(s)'} \left( \frac{C}{\lambda} \boldsymbol{I} + \boldsymbol{\delta}_{\boldsymbol{U}} \mathbf{H}^{(s)} \boldsymbol{A}_{\boldsymbol{U}}^{(s), -1} \mathbf{H}^{(s)'} \right)^{-1} \boldsymbol{\delta}_{\boldsymbol{U}} \mathbf{H}^{(s)}  \boldsymbol{A}_{\boldsymbol{U}}^{(s), -1} \\
& = \boldsymbol{A}_{\boldsymbol{U}}^{(s), -1} - \boldsymbol{\Delta}_{\boldsymbol{U}}^{(s)}  \boldsymbol{A}_{\boldsymbol{U}}^{(s), -1} ,
\end{split}
\end{equation}

where

\begin{equation}
\label{eq:delta_U}
\boldsymbol{\Delta}_{\boldsymbol{U}}^{(s)} \equiv \boldsymbol{A}_{\boldsymbol{U}}^{(s), -1} \boldsymbol{H}^{(s)'} \left( \frac{C}{\lambda} \boldsymbol{I} + \boldsymbol{\delta}_{\boldsymbol{U}} \mathbf{H}^{(s)} \boldsymbol{A}_{\boldsymbol{U}}^{(s), -1} \mathbf{H}^{(s)'} \right)^{-1} \boldsymbol{\delta}_{\boldsymbol{U}} \mathbf{H}^{(s)}.
\end{equation}

Similar result is obtained for Eq. \eqref{eq:t_V_j_s}. Because $U_j^{(s)} = \boldsymbol{A}_{\boldsymbol{U}}^{(s), -1} \boldsymbol{H}^{(s)} \boldsymbol{Y}_j$, using Eq. \eqref{eq:sherman-morrison} into Eq. \eqref{eq:t_U_j_s}, \eqref{eq:t_V_j_s} led us to achieve $\boldsymbol{U}^{(s)}$ and $\boldsymbol{V}^{(s)}$

\begin{eqnarray}
\boldsymbol{T}^{(s)} (\boldsymbol{U}) & = & \boldsymbol{U}^{(s)} - \boldsymbol{\Delta}_{\boldsymbol{U}}^{(s)} \boldsymbol{U}^{(s)},  \\
\boldsymbol{T}^{(s)} (\boldsymbol{V}) & = & \boldsymbol{V}^{(s)} - \boldsymbol{\Delta}_{\boldsymbol{V}}^{(s)}\boldsymbol{V}^{(s)}.
\end{eqnarray}

The distance $d^{(s)} (\boldsymbol{T}^{(s)} (\boldsymbol{U}), \boldsymbol{T}^{(s)} (\boldsymbol{V}))$ can be expressed as

\begin{equation}
\label{eq:distance_t}
\begin{split}
d^{(s)} (\boldsymbol{T}^{(s)} (\boldsymbol{U}), \boldsymbol{T}^{(s)} (\boldsymbol{V}^{(s)})) = & {\lVert \boldsymbol{U}^{(s)} - \boldsymbol{V}^{(s)} - \left( \boldsymbol{\Delta}_{\boldsymbol{U}}^{(s)} \boldsymbol{U}^{(s)} - \boldsymbol{\Delta}_{\boldsymbol{V}}^{(s)} \boldsymbol{V}^{(s)} \right)  \rVert}^2 = \\
 & { \lVert \boldsymbol{U}^{(s)} - \boldsymbol{V}^{(s)} \rVert}^2 + {\lVert \boldsymbol{\Delta}_{\boldsymbol{U}}^{(s)} \boldsymbol{U}^{(s)} - \boldsymbol{\Delta}_{\boldsymbol{V}}^{(s)} \boldsymbol{V}^{(s)} \rVert}^2 - \\
 & 2 { \lVert \boldsymbol{U}^{(s)} - \boldsymbol{V}^{(s)} \rVert} {\lVert \boldsymbol{\Delta}_{\boldsymbol{U}}^{(s)} \boldsymbol{U}^{(s)} - \boldsymbol{\Delta}_{\boldsymbol{V}}^{(s)} \boldsymbol{V}^{(s)} \rVert}.
\end{split}
\end{equation}

Since the solution $\boldsymbol{U}^{(s)} = \boldsymbol{V}^{(s)}$ is discarded, Eq. \eqref{eq:distance_t} can be divided by the distance $d^{(s)} (\boldsymbol{U}^{(s)}, \boldsymbol{V}^{(s)}) = { \lVert \boldsymbol{U}^{(s)} - \boldsymbol{V}^{(s)} \rVert}^2 > 0$,

\begin{equation}
\label{eq:distance_frac}
\begin{split}
\frac{d^{(s)} (\boldsymbol{T}^{(s)} (\boldsymbol{U}^{(s)}), \boldsymbol{T}^{(s)} (\boldsymbol{V}^{(s)}))}{d^{(s)} (\boldsymbol{U}^{(s)}, \boldsymbol{V}^{(s)})} = & 1 + {\frac{\lVert \boldsymbol{\Delta}_{\boldsymbol{U}}^{(s)} \boldsymbol{U}^{(s)} - \boldsymbol{\Delta}_{\boldsymbol{V}}^{(s)} \boldsymbol{V}^{(s)} \rVert}{ \lVert \boldsymbol{U}^{(s)} - \boldsymbol{V}^{(s)} \rVert}}^2 \\
& - 2 \frac{\lVert \boldsymbol{\Delta}_{\boldsymbol{U}}^{(s)} \boldsymbol{U}^{(s)} - \boldsymbol{\Delta}_{\boldsymbol{V}}^{(s)} \boldsymbol{V}^{(s)} \rVert}{ \lVert \boldsymbol{U}^{(s)} - \boldsymbol{V}^{(s)} \rVert} = \\
& \left( {\frac{\lVert \boldsymbol{\Delta}_{\boldsymbol{U}}^{(s)} \boldsymbol{U}^{(s)} - \boldsymbol{\Delta}_{\boldsymbol{V}}^{(s)} \boldsymbol{V}^{(s)} \rVert}{ \lVert \boldsymbol{U}^{(s)} - \boldsymbol{V}^{(s)} \rVert}} - 1 \right)^2,
\end{split}
\end{equation}

and applying Eq. \eqref{eq:ineq_distance},

\begin{equation}
\left( {\frac{\lVert \boldsymbol{\Delta}_{\boldsymbol{U}}^{(s)} \boldsymbol{U}^{(s)} - \boldsymbol{\Delta}_{\boldsymbol{V}}^{(s)} \boldsymbol{V}^{(s)} \rVert}{ \lVert \boldsymbol{U}^{(s)} - \boldsymbol{V}^{(s)} \rVert}} - 1 \right)^2 < q, \: q \in [0, 1).
\end{equation}

Because real terms powers to 2 are greater than 0, it is only needed to prove that

\begin{equation}
\label{eq:bound_distance}
\begin{split}
& \left( {\frac{\lVert \boldsymbol{\Delta}_{\boldsymbol{U}}^{(s)} \boldsymbol{U}^{(s)} - \boldsymbol{\Delta}_{\boldsymbol{V}}^{(s)} \boldsymbol{V}^{(s)} \rVert}{ \lVert \boldsymbol{U}^{(s)} - \boldsymbol{V}^{(s)} \rVert}} - 1 \right)^2 <  1, \\
-1 < & {\frac{\lVert \boldsymbol{\Delta}_{\boldsymbol{U}}^{(s)} \boldsymbol{U}^{(s)} - \boldsymbol{\Delta}_{\boldsymbol{V}}^{(s)} \boldsymbol{V}^{(s)} \rVert}{ \lVert \boldsymbol{U}^{(s)} - \boldsymbol{V}^{(s)} \rVert}} - 1 < 1, \\
0 < & {\frac{\lVert \boldsymbol{\Delta}_{\boldsymbol{U}}^{(s)} \boldsymbol{U}^{(s)} - \boldsymbol{\Delta}_{\boldsymbol{V}}^{(s)} \boldsymbol{V}^{(s)} \rVert}{ \lVert \boldsymbol{U}^{(s)} - \boldsymbol{V}^{(s)} \rVert}} <  2.
\end{split}
\end{equation}

Left inequality is assured, due to ${ \lVert \cdot \rVert}_2 \ge 0$. Powering the fraction to 2 and applying norm properties,

\begin{eqnarray}
\label{eq:triangle_ineq}
{ \lVert \boldsymbol{x} + \boldsymbol{y} \rVert} \le { \lVert \boldsymbol{x} \rVert} + { \lVert  \boldsymbol{y} \rVert}, \\
\label{eq:product}
{ \lVert \boldsymbol{B} \boldsymbol{C} \rVert} \le { \lVert \boldsymbol{B} \rVert} { \lVert  \boldsymbol{C} \rVert}, \\
\end{eqnarray}

we have

\begin{equation}
\label{eq:rayleigh}
\begin{split}
{\frac{{\lVert \boldsymbol{\Delta}_{\boldsymbol{U}}^{(s)} \boldsymbol{U}^{(s)} - \boldsymbol{\Delta}_{\boldsymbol{V}}^{(s)} \boldsymbol{V}^{(s)} \rVert}^2}{ {\lVert \boldsymbol{U}^{(s)} - \boldsymbol{V}^{(s)} \rVert}^2}} & \le {\frac{{\lVert \boldsymbol{\Delta}_{\boldsymbol{U}}^{(s)} \boldsymbol{U}^{(s)} \rVert}^2 + { \lVert \boldsymbol{\Delta}_{\boldsymbol{V}}^{(s)} \boldsymbol{V}^{(s)} \rVert}^2}{ {\lVert \boldsymbol{U}^{(s)} - \boldsymbol{V}^{(s)} \rVert}^2}} \\
&  \le {\frac{{\lVert \boldsymbol{\Delta}_{\boldsymbol{U}}^{(s)} \rVert}^2 {\lVert \boldsymbol{U}^{(s)} \rVert}^2 + { \lVert \boldsymbol{\Delta}_{\boldsymbol{V}}^{(s)} \rVert}^2 { \lVert \boldsymbol{V}^{(s)} \rVert}^2}{ {\lVert \boldsymbol{U}^{(s)} - \boldsymbol{V}^{(s)} \rVert}^2}} 
\end{split}
\end{equation}

and the problem of the maximum value of Eq. \eqref{eq:rayleigh} is the generalized Rayleigh quotient \cite{parlett1998the},

\begin{equation}
\label{eq:max_eigenvalue}
\max_{\boldsymbol{U}^{(s)}, \boldsymbol{V}^{(s)}} \frac{\begin{pmatrix} \boldsymbol{U}^{(s)} & \boldsymbol{V}^{(s)} \end{pmatrix} \begin{pmatrix} {\lVert \boldsymbol{\Delta}_{\boldsymbol{U}}^{(s)} \rVert}^2 & 0 \\ 0 & {\lVert \boldsymbol{\Delta}_{\boldsymbol{V}}^{(s)} \rVert}^2 \end{pmatrix} \begin{pmatrix} \boldsymbol{U}^{(s)} \\ \boldsymbol{V}^{(s)} \end{pmatrix}} {\begin{pmatrix} \boldsymbol{U}^{(s)} & \boldsymbol{V}^{(s)} \end{pmatrix} \begin{pmatrix} 1 & -1 \\ -1 & 1 \end{pmatrix} \begin{pmatrix} \boldsymbol{U}^{(s)} \\ \boldsymbol{V}^{(s)} \end{pmatrix}} = \max_{\boldsymbol{W}} \frac{\boldsymbol{W}' \boldsymbol{X} \boldsymbol{W}}{\boldsymbol{W}' \boldsymbol{Y} \boldsymbol{W}}.
\end{equation}

This problem is equivalent to

\begin{equation}
\label{eq:dual}
\begin{split}
& \max_{\boldsymbol{W}}  \: \boldsymbol{W}' \boldsymbol{X} \boldsymbol{W} \\
& \text{s.t.} \:  \boldsymbol{W}' \boldsymbol{Y} \boldsymbol{W} = K
\end{split}
\end{equation}

where $K$ in this problem is the distance between $\boldsymbol{U}^{(s)}, \boldsymbol{V}^{(s)}$, which is nonzero because that problem was discarded. This can be solved used Lagrange multipliers,

\begin{equation}
\label{eq:lagrange}
L_{\boldsymbol{W}} = \boldsymbol{W}' \boldsymbol{X} \boldsymbol{W} - \gamma \left(\boldsymbol{W}' \boldsymbol{Y} \boldsymbol{W} - K \right).
\end{equation}

Maximizing respect to $\boldsymbol{W}$, a Generalized Eigenvalue Problem (GEP) is obtained,

\begin{equation}
\label{eq:gep}
\begin{split}
\nabla L_{\boldsymbol{W}} = 2 \left( \boldsymbol{X} - \gamma \boldsymbol{Y} \right) \boldsymbol{W} = 0 \\
\boldsymbol{X}  \boldsymbol{W} = \gamma \boldsymbol{Y} \boldsymbol{W}.
\end{split}
\end{equation}

In the GEP, the eigenvalues can be calculated as $det(\boldsymbol{X} - \gamma \boldsymbol{Y}) = 0$

\begin{equation}
\label{eq:eigenvalue}
\begin{split}
det \begin{pmatrix}
{\lVert \boldsymbol{\Delta}_{\boldsymbol{U}}^{(s)} \rVert}^2 - \gamma & \gamma \\
\gamma & {\lVert \boldsymbol{\Delta}_{\boldsymbol{V}}^{(s)} \rVert}^2- \gamma
\end{pmatrix} & = {\lVert \boldsymbol{\Delta}_{\boldsymbol{U}}^{(s)} \rVert}^2 {\lVert \boldsymbol{\Delta}_{\boldsymbol{V}}^{(s)} \rVert}^2 - \gamma \left( {\lVert \boldsymbol{\Delta}_{\boldsymbol{U}}^{(s)} \rVert}^2 + {\lVert \boldsymbol{\Delta}_{\boldsymbol{V}}^{(s)} \rVert}^2 \right) = 0\\
\gamma & = \frac{{\lVert \boldsymbol{\Delta}_{\boldsymbol{U}}^{(s)} \rVert}^2 {\lVert \boldsymbol{\Delta}_{\boldsymbol{V}}^{(s)} \rVert}^2}{{\lVert \boldsymbol{\Delta}_{\boldsymbol{U}}^{(s)} \rVert}^2 + {\lVert \boldsymbol{\Delta}_{\boldsymbol{V}}^{(s)} \rVert}^2} =  \frac{1}{\frac{1}{{\lVert \boldsymbol{\Delta}_{\boldsymbol{U}}^{(s)} \rVert}^2} + \frac{1}{{\lVert \boldsymbol{\Delta}_{\boldsymbol{V}}^{(s)} \rVert}^2}}.
\end{split}
\end{equation}

$\gamma_{max}$ is the maximum eigenvalue of the quotient in Eq. \eqref{eq:max_eigenvalue}. From Eq. \eqref{eq:distance_frac} and \eqref{eq:bound_distance}, if $\gamma_{max} < 4$ then condition from Equation \eqref{eq:ineq_distance} is assured.

Using norm property in Eq. \eqref{eq:triangle_ineq}, and adding previous knowledge ${\lVert \boldsymbol{x} \rVert} < {\lVert \boldsymbol{y} \rVert}$, a bottom bound can be set. Taking inverse,
	
\begin{equation}
\label{eq:triangle_ineq_2}
\begin{split}
{ \lVert \boldsymbol{x} \rVert} - { \lVert  \boldsymbol{y} \rVert} \le { \lVert \boldsymbol{x} + \boldsymbol{y} \rVert} \le { \lVert \boldsymbol{x} \rVert} + { \lVert  \boldsymbol{y} \rVert}, \\
\frac{1}{ \lVert \boldsymbol{x} \rVert + { \lVert  \boldsymbol{y} \rVert}} \le \frac{1}{ \lVert \boldsymbol{x} + \boldsymbol{y} \rVert} \le \frac{1}{ \lVert \boldsymbol{x} \rVert - { \lVert  \boldsymbol{y} \rVert}}.
\end{split}
\end{equation}

If ${\lVert \boldsymbol{y} \rVert} < {\lVert \boldsymbol{x} \rVert}$, the same reasoning could be followed. From norm property in Eq. \eqref{eq:product}, an upper bound for norm inverse matrix can be set,

\begin{equation}
\label{eq:product_2}
\begin{split}
\lVert \boldsymbol{I} \rVert = \lVert \boldsymbol{B} \boldsymbol{B}^{-1} \rVert & \le \lVert \boldsymbol{B} \rVert \lVert \boldsymbol{B}^{-1} \rVert \le 1 \\
\lVert \boldsymbol{B}^{-1} \rVert & \le \frac{1}{\lVert \boldsymbol{B} \rVert}
\end{split}
\end{equation}

Applying Eq. \eqref{eq:triangle_ineq_2} and \eqref{eq:product_2} in the definition of $\boldsymbol{\Delta}_{\boldsymbol{U}}^{(s)}$ in \eqref{eq:delta_U}, an upper bound can be found, 

\begin{equation}
\label{eq:bound_delta}
{\lVert \boldsymbol{\Delta}_{\boldsymbol{U}}^{(s)} \rVert}^2 \le \frac{\alpha_{\boldsymbol{U}}^{(s)} {\lVert \boldsymbol{\delta}_{\boldsymbol{U}} \rVert}^2 }{\frac{C^2}{\lambda^2} - \alpha_{\boldsymbol{U}}^{(s)} {\lVert \boldsymbol{\delta}_{\boldsymbol{U}} \rVert}^2 }
\end{equation}

Where $\alpha_{\boldsymbol{U}}^{(s)} \equiv {\lVert \boldsymbol{A}_{\boldsymbol{U}}^{(s), -1} \rVert}^2 {\lVert \boldsymbol{H}^{(s)} \rVert}^4$. Replacing in Eq. \eqref{eq:eigenvalue}, it is reached to the inequality

\begin{equation}
\label{eq:gamma_max}
\gamma \le  \frac{\alpha_{\boldsymbol{U}}^{(s)} \alpha_{\boldsymbol{V}}^{(s)} { \lVert \boldsymbol{\delta}_{\boldsymbol{U}} \rVert }^2 { \lVert \boldsymbol{\delta}_{\boldsymbol{V}} \rVert }^2 }{\frac{C^2}{\lambda^2} ( \alpha_{\boldsymbol{U}}^{(s)} { \lVert \boldsymbol{\delta}_{\boldsymbol{U}} \rVert }^2 + \alpha_{\boldsymbol{V}}^{(s)} { \lVert \boldsymbol{\delta}_{\boldsymbol{V}} \rVert }^2) - 2 \alpha_{\boldsymbol{U}}^{(s)} \alpha_{\boldsymbol{V}}^{(s)} { \lVert \boldsymbol{\delta}_{\boldsymbol{U}} \rVert }^2 { \lVert \boldsymbol{\delta}_{\boldsymbol{V}} \rVert }^2}
\end{equation}

so $C$ and $\lambda$ can be imposed, the maximum eigenvalue to be under condition $\gamma_{max} < 4$ in equation  \eqref{eq:gamma_max},

\begin{equation}
\label{eq:lambda_cond}
\lambda_{max} < \frac{2 C}{3 } \sqrt{\frac{{ \alpha_{\boldsymbol{U}}^{(s)} \lVert \boldsymbol{\delta}_{\boldsymbol{U}} \rVert }^2 + \alpha_{\boldsymbol{V}}^{(s)} { \lVert \boldsymbol{\delta}_{\boldsymbol{V}} \rVert }^2}{ \alpha_{\boldsymbol{U}}^{(s)} \alpha_{\boldsymbol{V}}^{(s)} { \lVert \boldsymbol{\delta}_{\boldsymbol{U}} \rVert }^2 { \lVert \boldsymbol{\delta}_{\boldsymbol{V}} \rVert }^2}}.
\end{equation}

After consider $\eta^{(s)}$ as

\begin{equation}
\label{eq:eta}
\eta^{(s)} \equiv \frac{ \lVert \boldsymbol{A}_{\boldsymbol{U}}^{(s), -1} \rVert }{ \lVert \boldsymbol{A}_{\boldsymbol{V}}^{(s), -1} \rVert } =  \frac{ \lVert (  \frac{\mathbf{I}}{C} +\mathbf{H}^{(s)'}\mathbf{H}^{(s)} + \frac{\lambda}{C} \mathbf{H}^{(s)'} \hat{\boldsymbol{F}}_{\boldsymbol{U}} \hat{\boldsymbol{F}}_{\boldsymbol{U}}' \mathbf{H}^{(s)})^{-1}  \rVert }{ \lVert (  \frac{\mathbf{I}}{C} +\mathbf{H}^{(s)'}\mathbf{H}^{(s)} + \frac{\lambda}{C} \mathbf{H}^{(s)'} \hat{\boldsymbol{F}}_{\boldsymbol{V}} \hat{\boldsymbol{F}}_{\boldsymbol{V}}' \mathbf{H}^{(s)} )^{-1} \rVert }
\end{equation}

and replacing $\alpha_{\boldsymbol{U}}^{(s)}, \alpha_{\boldsymbol{V}}^{(s)}$ into Equation \eqref{eq:lambda_cond},

\begin{equation}
\label{eq:lambda_cond_2}
\lambda_{max} < \frac{2 C}{3 \lVert \boldsymbol{A}_{\boldsymbol{U}}^{(s), -1} \rVert } \sqrt{\frac{ \eta^{(s)} {  \lVert \boldsymbol{\delta}_{\boldsymbol{U}} \rVert }^2 +  { \lVert \boldsymbol{\delta}_{\boldsymbol{V}} \rVert }^2}{ { \lVert \boldsymbol{\delta}_{\boldsymbol{U}} \rVert }^2 { \lVert \boldsymbol{\delta}_{\boldsymbol{V}} \rVert }^2}} \equiv \lambda_{bound}.
\end{equation}

$\lambda_{bound} $ values can be obtained numerically, by finding the zero in the following equation

\begin{equation}
\label{eq:F}
H(\lambda) = \lambda - \frac{2 C}{3 \lVert \boldsymbol{A}_{\boldsymbol{U}}^{(s), -1} \rVert } \sqrt{\frac{ \eta^{(s)} {  \lVert \boldsymbol{\delta}_{\boldsymbol{U}} \rVert }^2 + { \lVert \boldsymbol{\delta}_{\boldsymbol{V}} \rVert }^2}{ { \lVert \boldsymbol{\delta}_{\boldsymbol{U}} \rVert }^2 { \lVert \boldsymbol{\delta}_{\boldsymbol{V}} \rVert }^2}},
\end{equation}

because it is an implicit equation, where $\lVert \boldsymbol{A}_{\boldsymbol{U}}^{(s), -1} \rVert$ and $\eta^{(s)}$ depends on $\lambda$. However, $\lambda_{bound}$ can be relaxed using norm property in Equation \eqref{eq:product_2},

\begin{equation*}
\begin{split}
	{ \lVert \boldsymbol{A}_{\boldsymbol{U}}^{(s), -1} \rVert } \le  \frac{ 1 }{ \lVert  \frac{\mathbf{I}}{C} +\mathbf{H}^{(s)'}\mathbf{H}^{(s)} \rVert - \frac{\lambda}{C} \lVert  \mathbf{H}^{(s)'} \hat{\boldsymbol{F}}_{\boldsymbol{U}} \hat{\boldsymbol{F}}_{\boldsymbol{U}}' \mathbf{H}^{(s)}  \rVert }, \\
\frac{ 1 }{ \lVert \boldsymbol{A}_{\boldsymbol{U}}^{(s), -1} \rVert } \ge  \lVert  \frac{\mathbf{I}}{C} +\mathbf{H}^{(s)'}\mathbf{H}^{(s)} \rVert - \frac{\lambda}{C} \lVert  \mathbf{H}^{(s)'} \hat{\boldsymbol{F}}_{\boldsymbol{U}} \hat{\boldsymbol{F}}_{\boldsymbol{U}}' \mathbf{H}^{(s)}  \rVert,
\end{split}
\end{equation*}

thus, a more restrictive bound $\lambda_{bound}'$ can be set,

\begin{equation}
\label{eq:lambda_cond_3}
\begin{split}
\lambda & < \frac{2 C}{3 } (\lVert  \frac{\mathbf{I}}{C} +\mathbf{H}^{(s)'}\mathbf{H}^{(s)} \rVert - \frac{\lambda}{C} \lVert  \mathbf{H}^{(s)'} \hat{\boldsymbol{F}}_{\boldsymbol{U}} \hat{\boldsymbol{F}}_{\boldsymbol{U}}' \mathbf{H}^{(s)}  \rVert) \sqrt{\frac{ \eta^{(s)} {  \lVert \boldsymbol{\delta}_{\boldsymbol{U}} \rVert }^2 +  { \lVert \boldsymbol{\delta}_{\boldsymbol{V}} \rVert }^2}{ { \lVert \boldsymbol{\delta}_{\boldsymbol{U}} \rVert }^2 { \lVert \boldsymbol{\delta}_{\boldsymbol{V}} \rVert }^2}}, \\
\lambda & < \frac{2 \lVert  \mathbf{I} + C \mathbf{H}^{(s)'}\mathbf{H}^{(s)} \rVert} {3 (1 + \lVert  \mathbf{H}^{(s)'} \hat{\boldsymbol{F}}_{\boldsymbol{U}} \hat{\boldsymbol{F}}_{\boldsymbol{U}}' \mathbf{H}^{(s)}  \rVert)} \sqrt{\frac{ \eta^{(s)} {  \lVert \boldsymbol{\delta}_{\boldsymbol{U}} \rVert }^2 +  { \lVert \boldsymbol{\delta}_{\boldsymbol{V}} \rVert }^2}{ { \lVert \boldsymbol{\delta}_{\boldsymbol{U}} \rVert }^2 { \lVert \boldsymbol{\delta}_{\boldsymbol{V}} \rVert }^2}} \equiv \lambda_{bound}'. \\
\end{split}
\end{equation}

It is trivial to see that, if $\lambda \in (0, \lambda_{bound}')$, then $\lambda \in (0, \lambda_{bound})$. Although $\lambda$ is still implicit in Equation \eqref{eq:lambda_cond_3} through $\eta^{(s)}$, in the same Section this problem could be avoided.

\section{Discussion}
\label{sec:discussion}

\subsection{$\lambda$ condition}
\label{subsec:lambda}

For $\lambda$ values that assures the condition from Eq. \eqref{eq:lambda_cond_3}, then the inequality in Eq. \eqref{eq:ineq_distance} is also assured. Eq. \eqref{eq:ineq_distance} is much restrictive than condition from Banach fixed-point theorem, 

\begin{equation*}
d(\boldsymbol{T} (\boldsymbol{U}), \boldsymbol{T} (\boldsymbol{V})) \le q d(\boldsymbol{U}, \boldsymbol{V}), \: q \in [0, 1),
\end{equation*}

which means that, under certain condition of $\lambda$, there is an upper bound that allows to formulate that NCELM as a fixed-point iteration. Moreover, because sequence Eq. \eqref{eq:solution_iter} is a fixed-point iteration, $\lim_{r \to\infty} \boldsymbol{B}_{j, (r)} = \boldsymbol{B}_j^{*}$, with

\begin{equation} \label{eq:B_solution}
\boldsymbol{B}_j^{*} =
\begin{pmatrix}
\boldsymbol{\beta}_{j}^{(1)*} \\
\vdots \\
\boldsymbol{\beta}_{j}^{(s)*} \\
\vdots \\
\boldsymbol{\beta}_{j}^{(S)*} \\
\end{pmatrix},
\end{equation}

the solution of the system, $lim_{r \to\infty} \boldsymbol{F}_{j, (r)} = \boldsymbol{F}_j^{*}$, thus by definition of $\boldsymbol{\delta}_{(r)}$,

\begin{equation}
\label{eq:delta}
\boldsymbol{\delta}_{(r)} = {\boldsymbol{F}_{j, (r)}} {\boldsymbol{F}_{j, (r)}}' - {\boldsymbol{F}_{j, (r - 1)}} {\boldsymbol{F}_{j, (r - 1)}}' \rightarrow \boldsymbol{0},
\end{equation}

 and condition for $\lambda$ in Eq. \eqref{eq:lambda_cond} is relaxed over the iterations, since the upper bound for $\lambda$ increases,

\begin{equation}
\label{eq:lambda_cond_r}
\lim_{r \to\infty} \sqrt{\frac{{ \eta_{(r)}^{(s)} \lVert \boldsymbol{\delta}_{(r)} \rVert }^2 + { \lVert \boldsymbol{\delta}_{(r-1)} \rVert }^2}{{ \lVert \boldsymbol{\delta}_{(r)} \rVert }^2 { \lVert \boldsymbol{\delta}_{(r-1)} \rVert }^2}} = + \infty
\end{equation}

as long as $ \eta_{(r)}^{(s)} < + \infty$. And this is also assured, since matrix $\boldsymbol{A}_{(r - 1)}^{(s), -1}$ and $\boldsymbol{A}_{(r)}^{(s), -1}$ exist and are non singular, because of Equations \eqref{eq:U_j_s} and \eqref{eq:V_j_s}, so from Eq. \eqref{eq:eta},

\begin{equation*}
0 < \lVert \boldsymbol{A}_{(r-1)}^{(s), -1} \rVert < + \infty, \: \: 0 < \lVert \boldsymbol{A}_{(r)}^{(s), -1} \rVert < + \infty.
\end{equation*}

Eq. \eqref{eq:lambda_cond_r} implies that any $\lambda$ value can be chosen, whether $\lambda < \lambda_{bound}'$ or not, because the condition is relaxed over iterations and $\lambda_{bound}'$ becomes more and more large. If Eq. \eqref{eq:contraction} would be not fulfilled in the first iteration for $\lambda$, then $\hat{\lambda}< \lambda_{bound}'$ could be chosen, and the boundary would be relaxed during the training stage, until $\lambda < \lambda_{bound}'$. Using the base learners obtained at this point of the training stage, the fixed-point iteration could continue with $\lambda$.

\subsection{Experimental results}
\label{subsec:experiments}

Because the base learners converge to an ensemble optimum, the difference between the coefficient vectors in iteration $r$, and the values in the next iteration $r + 1$, always decreases. For the explanatory purpose, this Section shows graphically an example of this convergence\footnote{using NCELM implementation from \textit{pyridge}, https://github.com/cperales/pyridge}. The dataset \textit{qsar-biodegradation} from original paper experimental framework \cite{perales2020negative} is chosen,

\begin{table}[htpb]
	\centering
	\scriptsize
	\begin{tabular}{lrrrr}
		Dataset  &  Size  &  \#Attr.  & \#Classes  & {Class distribution} \\
		\hline
		\hline
		qsar-biodegradation	&	1055	&	41	&	2	&	(699, 356) \\
		\hline
\hline
\end{tabular}
\caption{Characteristics of the dataset \textit{qsar-biodegradation}: the number of patterns (\textit{Size}), attributes ($\#$\textit{Attr.}), classes ($\#$\textit{Classes}) and the distribution of instances within classes (\textit{Class distribution}).}
\label{table:dataset}
\end{table}%

Hyper-parameters are $C = 1$, $iterations = 10$ and $\lambda = \{10^{-6}, 10^{-5}, 10^{-4}\}$. To reduce the computational burden, vector norm chosen for plotting where not $L_2$ norm but $L_1$,

\begin{equation*}
	d(\boldsymbol{\beta_{j, (r)}}^{(s)}, \boldsymbol{\beta_{j, (r + 1)}}^{(s)}) = \lvert \boldsymbol{\beta_{j, (r + 1)}}^{(s)} - \boldsymbol{\beta_{j, (r)}}^{(s)} \rvert
\end{equation*}

since all the vector norms are equivalents \cite{arendt2009equivalent}, so all the norms of this difference show that it reduces to $0$. This norm is sum $\forall s, \: s = 1, \ldots, S$ and $\forall j, \: j = 1, \ldots, J$. 

\begin{figure}[H]
	\centering
	\label{fig:lambda_norm}
	\includegraphics[scale=0.5]{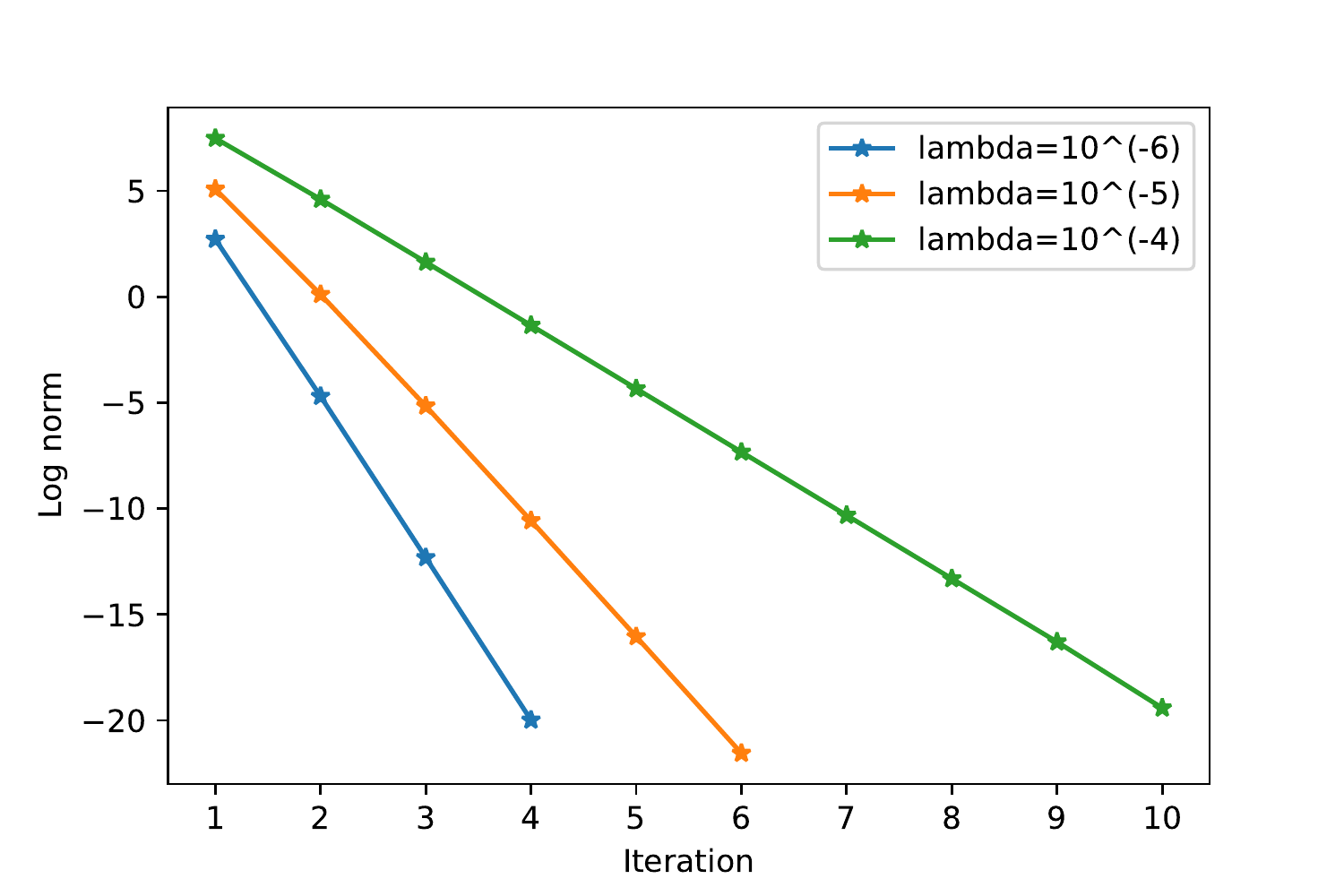}
	\caption{Norm of between the iterations, compared with different values of $\lambda$}
\end{figure}

\figurename{\ref{fig:lambda_norm}} shows how the difference decrease with the iterations\footnote{The code for this graphic is published online in a Jupyter notebook, \url{https://github.com/cperales/pyridge/blob/ncelm/NCELM_convergence.ipynb}}. The smaller the $\lambda$ is chosen, the faster the norm of the difference decreases to $0$. This means that the optimal convergence for $\ll \lambda$ is achieved faster than for high values, closer to the $\lambda$ boundaries.

\section{Conclusions}
\label{sec:conclusions}

In this paper, a new conceptualization of ensemble training has been presented for the Negative Correlation Extreme Learning Machine (NCELM) model. The training stage can be reformulated as a contraction mapping, and the update of the base learners is seen as a fixed-point iteration. Using Banach theorem, it was proved that the update of the base learners converge to a global optimum, in which error, regularization, and diversity terms from NCELM are optimized altogether.

Besides, this work highlight that reformulation of the training stage as a fixed-point iteration could help in the study of ensemble learning convergence. The re-conceptualization of base learner update in the ensemble as contraction mapping, along with Banach theorem, proves the convergence of the methods, and it is an interesting idea to explore in other ensemble learning methods.

%
\section*{Conflict of interest}

The authors declare that they have no conflict of interest.

\bibliographystyle{spmpsci}      
\bibliography{references}   

\end{document}